\documentclass[11pt,a4paper]{article}
\usepackage[hyperref]{naaclhlt2019}
\usepackage{times}
\usepackage{latexsym}
\usepackage{amssymb}
\usepackage{subfig}
\usepackage{multicol}
\usepackage{lipsum}
\usepackage{pifont}
\usepackage{array}
\usepackage{booktabs}
\usepackage{multirow}
\usepackage{graphicx}
\usepackage{xcolor}
\usepackage{amssymb}
\usepackage{float}
\usepackage{amsxtra}
\usepackage{times}
\usepackage{latexsym}
\usepackage{graphicx}
\usepackage{url}
\usepackage{multirow}
\usepackage{soul}
\newcommand{\argmax}{\operatornamewithlimits{argmax}}
\aclfinalcopy % Uncomment this line for the final submission
\def\aclpaperid{1579} %  Enter the acl Paper ID here

\title{Cross-topic distributional semantic representations via unsupervised mappings}

\author{ 
	Eleftheria Briakou$^{1,2}$\thanks{~\ The research was performed when the author was an undergraduate researcher at School of ECE, NTUA in Athens, Greece.}, \ Nikos Athanasiou$^{2}$, Alexandros Potamianos$^{2,3}$ \\\\
	$^1$University of Maryland, College Park, MD\\
	$^2$School of ECE, National Technical University of Athens, Athens, Greece \\
	$^3$Signal Analysis and Interpretation Laboratory (SAIL), USC, Los Angeles, USA\\  
    {\tt}\\
	{\tt ebriakou@cs.umd.edu, athn.nik@gmail.com, potam@central.ntua.gr}\\
	}
\date{}

\begin{document}
\maketitle
\begin{abstract}
In traditional Distributional Semantic Models (DSMs) the multiple senses of a polysemous word are conflated into a single vector space representation. In this work, we propose a DSM that learns multiple distributional representations of a word based on different topics. First, a separate DSM is trained for each topic and then each of the topic-based DSMs is aligned to a common vector space. Our unsupervised mapping approach is motivated by the hypothesis that words preserving their relative distances in different topic semantic sub-spaces constitute robust \textit{semantic anchors} that define the mappings between them. Aligned cross-topic representations achieve state-of-the-art results for the task of contextual word similarity. Furthermore, evaluation on NLP downstream tasks shows that multiple topic-based embeddings outperform single-prototype models.

\end{abstract}

\section{Introduction}
Word-level representation learning algorithms adopt the \textit{distributional hypothesis} \cite{harris}, presuming a correlation between the distributional and the semantic relationships of words. Typically, these models encode the contextual information of words into dense feature vectors---often referred to as \textit{embeddings}---of a $k$-dimensional space, thus creating a Vector Space Model (VSM) of lexical semantics. Such embeddings have been successfully applied to various natural language processing applications, including information retrieval \cite{Manning2008IntroductionIntroduction}, sentiment analysis \cite{sent}, and machine translation \cite{amiri2016learning,umd}.   

Despite their popularity, traditional DSMs rely solely on models where each word is uniquely represented by one point in the vector space. From a linguistic perspective, these models cannot capture the distinct meanings of polysemous words (e.g., \textit{bank} or \textit{cancer}), resulting in conflated word representations of diverse contextual semantics. 

To alleviate this problem, DSMs with multiple representations per word have been proposed in the literature, based on clustering local contexts of individual words~\cite{reisinger2010multi,tian2014probabilistic,neelakantan2014efficient}. An alternative way to train multiple representation DSMs is to utilize semantic lexical resources \cite{rothe2015autoextend,pilehvar2016conflated}. \citet{Christopoulou2018MixtureModels}, based on the assumption that typically words appear with a specific sense in each topic, proposed a topic-based semantic mixture model that exploits a combination of similarities estimated on topic-based DSMs for the computation of semantic similarity between words. Their model performs well for a variety of semantic similarity tasks; however, it lacks a unified representation of multiple senses in a common semantic space. The problem of defining transformations between embeddings---trained independently under different corpora---has been previously examined in various works, such as machine translation \cite{Mikolov2013LinguisticRepresentations,Xing2015NormalizedWE,artetxe2016learning}, induction of historical embeddings \cite{hamilton2016diachronic} and lexical resources enrichment \cite{Prokhorov2017LearningBridging}.

Following this line of research, we induce the creation of multiple cross-topic word embeddings by projecting the semantic representations of topic-based DSMs to a unified semantic space. We investigate different ways to perform the mappings from the topic sub-spaces to the unified semantic space, and propose a completely unsupervised approach to extract \textit{semantic anchors} that define those mappings. Furthermore, we claim that polysemous words change their meaning in different topic domains; this is reflected in relative shifts of their distributional representations in different topic-based DSMs. On the other hand, semantic anchors should have consistent semantic relationships regardless of the domain they reside in. Hence, their distributions of similarity values should also be similar across different domains. Finally, we apply a smoothing technique to each word's set of topic embeddings, resulting in representations with fine-grained semantics. 

To our knowledge, this is the first time that mappings between semantic spaces are applied to the problem of learning multiple embeddings for polysemous words. Our multi-topic word representations are evaluated on the contextual semantic similarity task and yield state-of-the-art performance compared to other unsupervised multi-prototype word embedding approaches. We further perform experiments on two NLP downstream tasks: text classification and paraphrase identification and demonstrate that our learned word representations consistently provide higher performance 
than single-prototype word embedding models. The code of the present work is publicly available\footnote{\url{https://github.com/Elbria/utdsm_naacl2018}}.
 
\section{Related Work}

Methods that assign multiple distributed representations per word can be grouped into two broad categories.\footnote{We limit our discussion to related works that use monolingual DSMs and corpora.} Unsupervised methods induce multiple word representations without leveraging semantic lexical resources. \citet{reisinger2010multi} were the first to create a multi-prototype DSM with a fixed number of vectors assigned to each word. In their model, the centroids of context-dependent clusters were used to create a set of  ``sense-specific'' vectors for each target word. Based on similar clustering approaches, follow-up works introduced neural network architectures that incorporated both local and global context in a joint training objective~\cite{huang2012improvin}, as well as methods that jointly performed word sense clustering and embedding learning as in  \citet{neelakantan2014efficient,li2015multi}. A probabilistic framework was introduced by \citet{tian2014probabilistic}, where the Skip-Gram model of Word2Vec was modified to learn multiple embedding vectors. Furthermore, latent topics were integrated into the Skip-Gram model, resulting in topical word embeddings which modeled the semantics of a word under different contexts \cite{liu2015topical,Liu2015cont,Nguyen2017AMM}. Another topic-related embedding creation approach was proposed in \citet{Christopoulou2018MixtureModels} where a mixture of topic-based semantic models was extracted by topical adaptation of in-domain corpora. Other approaches used autoencoders \cite{amiri2016learning}, convolutional neural networks designed to produce context representations that reflected the order of words in a context \cite{zheng2017learning} and reinforcement learning \cite{Lee2017MUSEMU,gr}.

Supervised approaches, based on prior knowledge acquired by sense inventories (e.g., WordNet) along with word sense disambiguation algorithms, were also introduced for sense-specific representations extraction \cite{chen2014unified, iacobacci2015sensembed}. In other works,   pre-trained word embeddings have been extended to embeddings of lexemes and synsets \cite{rothe2015autoextend} or were de-conflated into their constituent sense representations \cite{pilehvar2016conflated} by exploiting semantic lexical resources.

\section{Unified multi-Topic DSM (UTDSM)}

Our system follows a four-step approach:
\begin{enumerate}
\item \textbf{Global Distributional Semantic Model.}
Given a large collection of text data we train a DSM that encodes the contextual semantics of each word into a single representation, also referred to as Global-DSM. 

\item \textbf{Topic-based Distributional Semantic Models.} Next, a topic model is trained using the same corpus. The topic model splits the corpus into $K$ (possibly overlapping) sub-corpora. A DSM is then trained from each sub-corpus resulting in  $K$ topic-based DSMs (TDSMs). The topical adaptation of the semantic space takes into account the contextual variations a word exhibits under different thematic domains and therefore leads to the creation of ``topic-specific'' vectors (topic embeddings).

\item \textbf{Mappings of topic embeddings.} Next, we map the vector space of each topic-based DSM to the shared space of the Global-DSM, using a list of anchor words selected through an unsupervised self-learning scheme. In the unified semantic space each word is represented by a set of topic embeddings that were previously isolated in distinct vector spaces, thus creating a Unified multi-Topic DSM (UTDSM).

\item \textbf{Smoothing of topic embeddings.} 
As an optional step, we employ a smoothing approach in order to cluster a word's topic embeddings into $N$ Gaussian distributions via a Gaussian Mixture Model (GMM). This step lessens the noise introduced to our system through the semantic mappings and sparse training data. 
\end{enumerate}

\begin{figure}[!ht]
	\centering
	\includegraphics[scale=0.33]{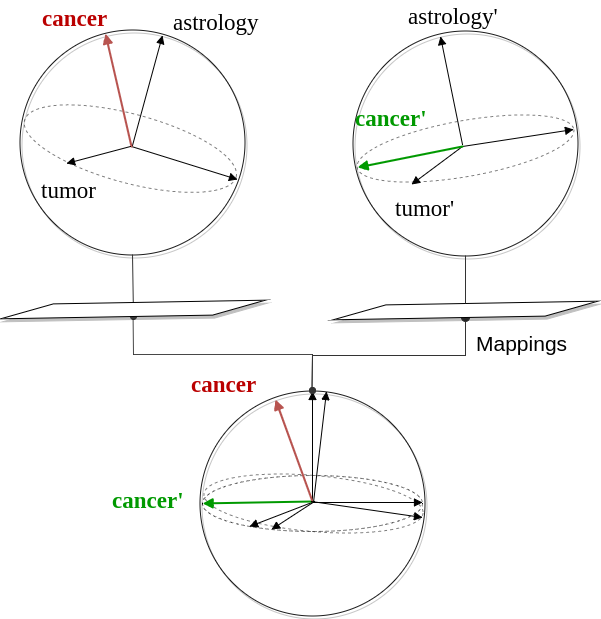}
    \captionsetup{font=small}
    \caption{Simplified depiction summarizing the intuition behind the alignment process of topic embeddings. In the unified vector space, the polysemous word \textit{cancer} is represented by two topic vectors that capture different semantic properties of the word under a zodiacal and a medical topic. Words \textit{astrology} and \textit{tumor} are examples of \textit{semantic anchors} that define the mappings.
    }
    \label{multi}
\end{figure}

\begin{figure*}[!ht]
\minipage{0.25\textwidth}
  \includegraphics[width=\linewidth]{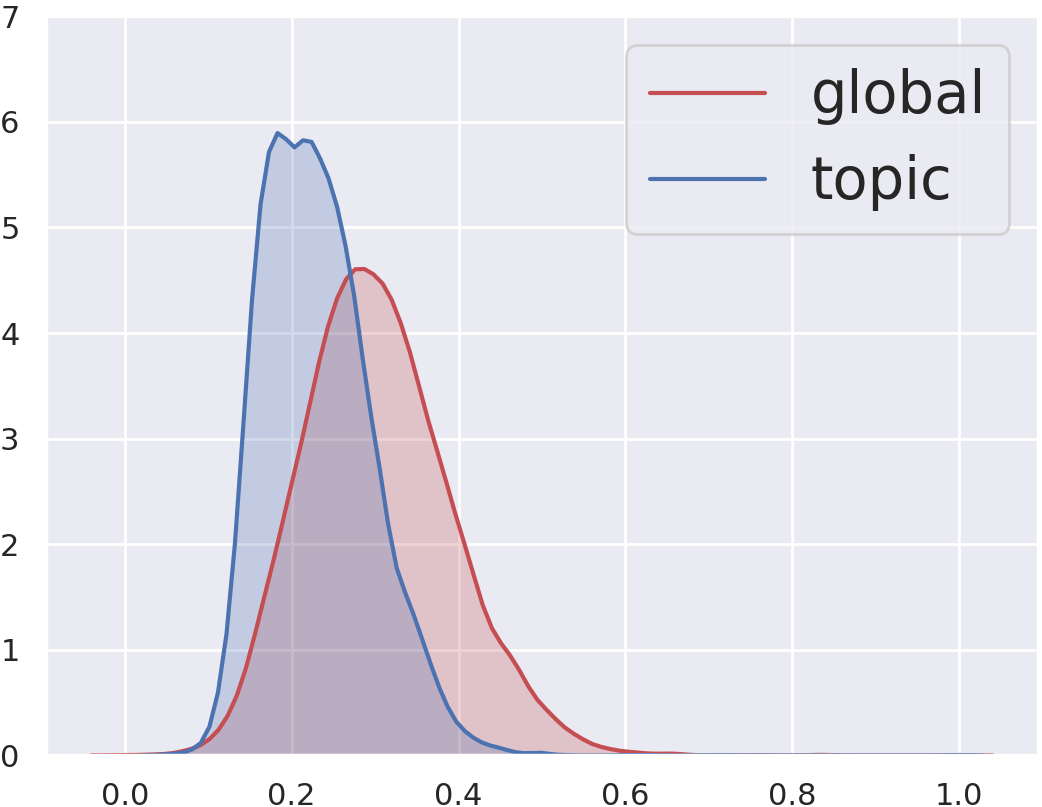}
\caption*{view} 

\endminipage\hfill
\minipage{0.24\textwidth}
  \includegraphics[width=\linewidth]{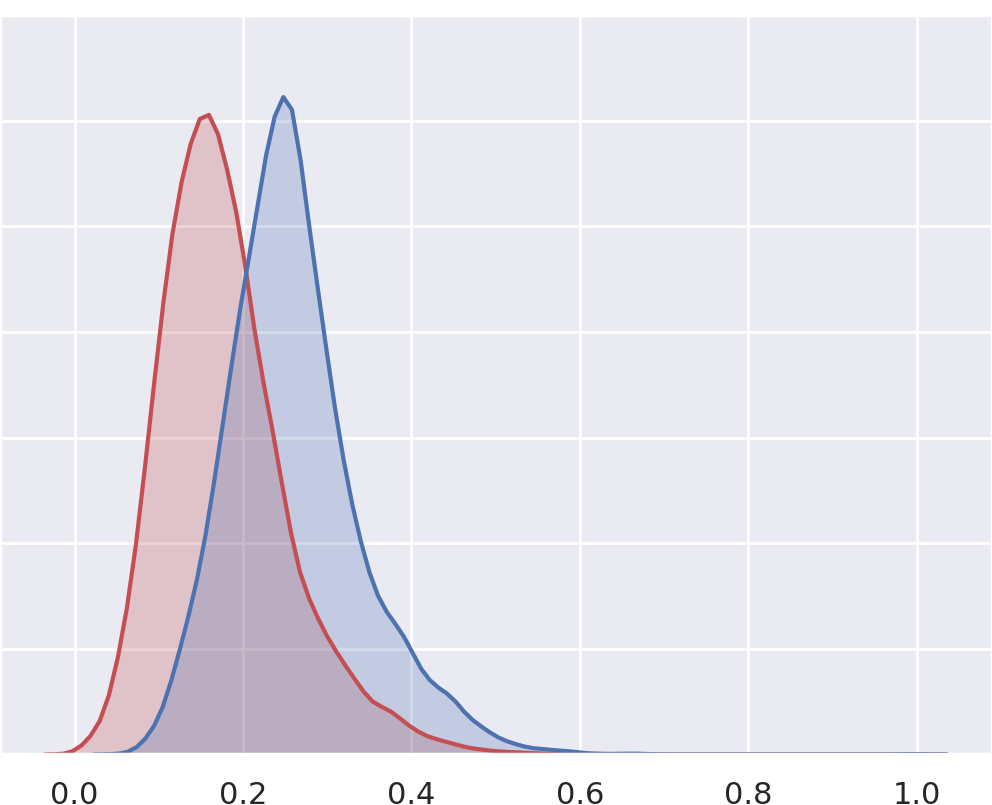}
 \caption*{crater} 

\endminipage\hfill
\minipage{0.24\textwidth}%
  \includegraphics[width=\linewidth]{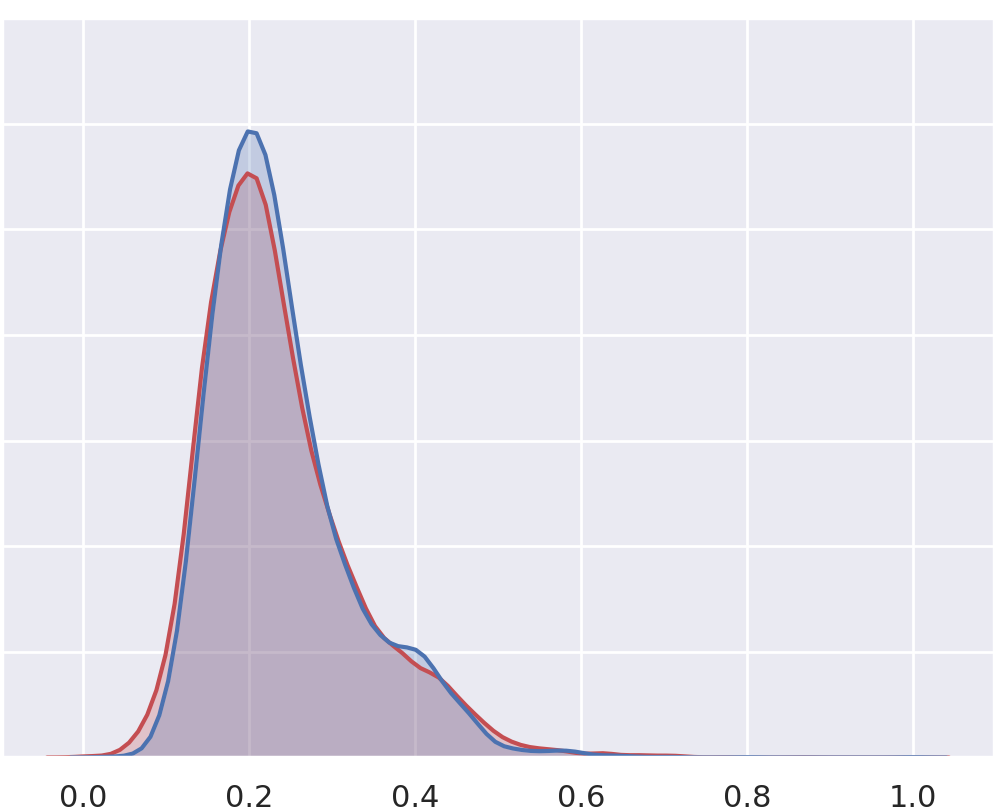}
  \caption*{professor} 

\endminipage\hfill
\minipage{0.24\textwidth}%
  \includegraphics[width=\linewidth]{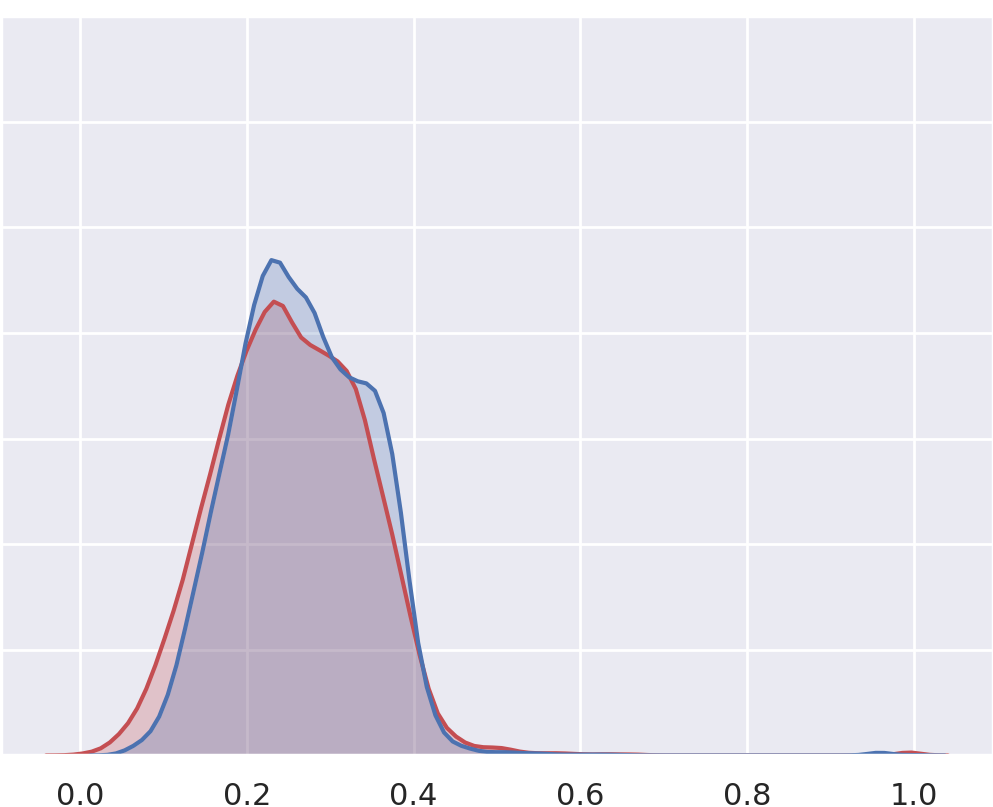}
  \caption*{october} 
\endminipage
\caption{Similarity distributions of four different words (corresponding to the smoothed density estimates of the similarity matrices) in topic domain space as defined in Equation~\ref{eq:average_topic_distr} and global space $s_g^i$. Selected anchors (``professor'' and ``october'') have more similar distributions in the global and topic spaces, when compared to unselected ones (``view'' and ``crater''). We observe that the selected anchors are less ambiguous, while the not selected ones are expected to have diverse contextual semantics.}
\label{fig:distributions}
\end{figure*}

\subsection{Topic-based Distributional Semantic Models}
The first step towards the thematic adaptation of the semantic space is the induction of in-domain corpora, using the Latent Dirichlet Algorithm (LDA) \cite{blei2003latent}. LDA is a generative probabilistic model of a corpus. Its core idea is that documents are represented as random mixtures over topics; where each topic is defined as a probability distribution over a collection of words.
Given as input a corpus of documents, LDA trains a topic model and creates a distribution of words for each topic in the corpus. Using the trained LDA model, we infer a topic distribution for each sentence in the corpus. Afterward, following a soft clustering scheme each sentence is included in a topic-specific corpus when the posterior probability for the corresponding topic exceeds a predefined threshold. 
The resulting topic sub-corpora are then used to train
topic-based DSMs. Any of the DSM training algorithms proposed in the literature can be used for this purpose; in this paper, we opt for the Word2Vec model \cite{mikolov2013efficient}.

\subsection{Mappings of topic embeddings}\label{subseq:mappings}
The intrinsic non-determinism of the Word2Vec algorithm leads to the creation of continuous vector spaces that are not naturally aligned to a unified semantic reference space, precluding the comparison between words of different thematic domains. To circumvent this limitation, we need to map the word representations of TDSMs to a shared vector space. In particular, we hypothesize that TDSMs capture meaningful variations in usage of polysemous words, while the relative semantic distance between monosemous words is preserved.
This hypothesis motivated us to think of monosemous words as \textit{anchors} between semantic spaces, as illustrated in Figure \ref{multi}. One way to retrieve the list of anchors is to extract monosemous words from lexical resources such as WordNet \cite{Prokhorov2017LearningBridging}. However, this method is restricted to languages where such lexical resources exist and depends on the lexical coverage and quality of such resources.

To overcome the above limitations, we propose a fully unsupervised method for semantic anchor induction. Although the embeddings of the topic and global semantic vector spaces are not aligned, their corresponding similarity matrices (once normalized) are. Based on this observation, we compute the similarity between a given word and every other word in the vocabulary (similarity distribution) for the different topic and global spaces.  
Then, we assume that good semantic anchors should have similar similarity distributions across the topic-specific and the global space, as illustrated in Figure \ref{fig:distributions}. \citet{artetxe2018acl} was based on a  similar observation to align vector semantic spaces in bilingual machine translation context.

Let $V$ be the intersection of the Global-DSM and the $K$ TDSMs vocabularies and $d$ the embedding dimension. We then define $X_k \in \mathbb{R}^{|V| \times d}$ as the embedding matrix of the $k$-th TDSM, and $Y \in \mathbb{R}^{|V| \times d}$ as the embedding matrix of the global DSM, where the $i$-th row of each matrix corresponds to the unit normalized representation of a word in $V$. Then, we define $S_{k} = X_{k}X_{k}^T$, $S_{g} = YY^T$  $\in \mathbb{R}^{|V| \times |V|}$ to be the similarity distribution matrices for the $k$-th TDSM and the global-DSM, respectively. Then our objective is to extract a list of semantic anchors $A$ that minimizes the Euclidean distance between the two different similarity distributions. Specifically, for every word $i$ we calculate the average semantic distribution across all topics:
\begin{equation}
<\!s_k^{i}\!>_k \;= \;\frac{1}{K} \sum_{k=1}^{K} s_{k}^{i}\\
 \label{eq:average_topic_distr}
\end{equation}
\vspace{-0.4cm}
\begin{equation}
 \| <\!s_k^{i}\!>_k \; - \; s_{g}^{i} \; \|_2\;, \;\ \ \forall \; i = 1, \ldots, |V|
 \label{eq:anchors_selection}
\end{equation}
where $s_g^i$, $s_k^{i}$ is the $i$-th row of the $S_g$ and $S_k$ similarity matrix, respectively, representing the similarity distribution between word $i$ and every other word in the vocabulary $V$. 
We then choose $|A|$ anchors as the words with the smallest values according to criterion \ref{eq:anchors_selection}.
Furthermore,  we assume that there exists an orthogonal transformation matrix between the topic embeddings of the extracted semantic anchors of each TDSM (source space) and the corresponding representations of the global-DSM (target space). The orthogonality constraint on the transformation matrix is widely adopted by the literature for various semantic space alignment tasks \citep{Xing2015NormalizedWE,artetxe2016learning,ham}. 
Assume $\alpha_{k}^{j} \in \mathbb{R}^d$ is the vector representation of the $j$-th \textit{anchor} word in the source space and $\alpha_{g}^{j} \in \mathbb{R}^d$ is its corresponding vector representation in the target space. The transformation matrix $M_k \in \mathbb{R}^{d \times d}$ that projects the first space to the latter is learned via solving the following constraint optimization problem:\footnote{This problem is known as the orthogonal Procrustes problem and it has a closed form solution as proposed in \cite{Schonemann1966AProblem}.}

\begin{equation}
 \min_{M_k} \sum_{j=1}^{|A|}  \| M_k\alpha_k^j - \alpha_g^j\|^2_2, \ \ \text{s.t.} \ \  M_k M_k^T = \mathbb{I}
 \label{eq:map}
\end{equation}
The induction of multiple topic embeddings in the unified vector space is achieved via  applying Equation \ref{eq:map} to each TDSM.  Specifically, given a word and its $k$-th topic distributed representation $x_k \in \mathbb{R}^d$, we compute its projected representation $x'_k \in \mathbb{R}^d$ as follows:

\begin{equation}
 x'_k = M_k x_k
\end{equation}
 
\subsection{Smoothing of topic embeddings}
Starting from the set of aligned topic embeddings $\{ x'_k \}_{k=1}^{K}$ for each word, we learn a Gaussian Mixture Model with $N$ components, where closely positioned topic embeddings are assigned to the same component. This step operates as an implicit way of segmenting the space of topic embeddings for each word in order to capture more useful hyper-topics---union of topics---which better represent their different meanings. We suggest that each Gaussian distribution forms a semantically coherent unit that corresponds to closely related semantics of the target word. Subsequently, the mean vector of each Gaussian distribution is used as a representative vector of each component, leading to a new set of \textit{smoothed} topic embeddings $\{ x_n^*\}_{n=1}^{N}$ for each word, where $x^*_n \in \mathbb{R}^d$.

\section{Experimental Setup}

\subsection{DSM settings}
As our initial corpus we  used  the  English  Wikipedia,  containing $8.5$ million  articles~\cite{turney2012domain}. During the training of the topic model, we used the articles found in the Wikipedia corpus and employed the Gensim implementation of LDA~\cite{Rubenstein} setting the number of topics $K$ to $50$. Using a threshold of $0.1$, we followed a soft-clustering approach, to bootstrap the creation of topic sub-corpora, using our trained topic model. Finally, we used Gensim's implementation of Word2Vec and Continuous Bag-of-Words method to train both the global-DSM and the TDSMs. The context window parameter of Word2Vec is set to $5$, while the dimensionality $d$ of all the constructed DSMs is equal to $300$ or $500$.\footnote{Any parameter not mentioned is set to default values of the corresponding implementations (e.g., Word2Vec, Gensim LDA).}

\subsection{Semantic Anchors}

The number of \textit{semantic anchors} that determine the mappings between our source and target spaces is set to $|A|=5\,000$ \footnote{ We have experimented with different values of anchors from $\{1\,000,2\,000,3\,000,4\,000,5\,000\}$ and report results for the best setup.} according to our unsupervised approach (criterion~\ref{eq:anchors_selection}). Those are selected from 
the common set of words that are represented in all semantic spaces with $|V|\sim12\,000$.

As a second experiment, we randomly sample $|A|$ words from the vocabulary of each TDSM to define its transformation matrix. We repeat this experiment $10$ times, every time sampling a different list from the corresponding vocabulary and report average performance results.

\subsection{Gaussian Mixture Model}
To apply the smoothing technique on the set of a word's topic embeddings we use the Scikit-learn implementation of Gaussian Mixture Model clustering algorithm~\cite{scikit-learn}. We initialize the mean vector of each component using k-means algorithm and the parameters of the model are estimated using Expectation-Maximization (EM) algorithm.

\subsection{Contextual Semantic Similarity}
To estimate the semantic similarity between a pair of words provided in sentential context, we use the standard evaluation Stanford Contextual Word Similarity (SCWS)~\cite{huang2012improvin} dataset which consists of $2\,003$ word-pairs with assigned semantic similarity scores computed as the average estimations of several human annotators. Following the evaluation guidelines proposed in literature, we employ the $\mathrm{AvgSimC}$ and $\mathrm{MaxSimC}$ contextual metrics, firstly discussed in \citet{reisinger2010multi}. In particular, given the word-pair $(w, w')$, and their provided contexts $(c$, $c')$ we define:
\makeatletter
    \def\tagform@#1{\maketag@@@{\normalsize(#1)\@@italiccorr}}
\makeatother
\begin{equation}
    \small
    \begin{split} 
        &\mathrm{AvgSimC}(w, w') = \\
&\frac{1}{K^2}\sum\limits_{j=1}^{K} \sum\limits_{k=1}^{K} \mathrm{p}(j|w,c) \mathrm{p}(k|w',c') \mathrm{d}(x'_j(w), x'_k(w')),
    \end{split}
    \label{avgsim}
\end{equation}
\begin{equation}\label{maxsim}
    \small
    \begin{split} 
    \hspace{-2.1cm}
        \mathrm{MaxSimC}(w, w') = \mathrm{d}(\hat{x'}(w), \hat{x'}(w')),
    \end{split}
\end{equation}
Following the notation used in \ref{subseq:mappings}, $K$ is the number of topics returned by the trained LDA model,  $x'_j$ is the word embedding trained on the sub-corpus corresponding to the $j$-th topic after being projected to the unified vector space, $\mathrm{p}(j|w,c)$ denotes the posterior probability of topic $j$ returned by LDA given as input the context $c$ of word $w$, $\mathrm{d}$ denotes the cosine similarity between the two input representations and finally $\hat{x'}(w) = u_{\argmax_{1 \leq j \leq K} \mathrm{p}(j|w,c)}(w)$ is the vector representation of word $w$ that corresponds to the topic with the maximum posterior for $c$. Intuitively, a higher score in $\mathrm{MaxSimC}$ indicates the existence of more robust multi-topic word representations. On the other hand, $\mathrm{AvgSimC}$ provides a topic-based smoothed result across different embeddings.

\subsection{Downstream NLP Tasks}
Besides the standard evaluation benchmark of contextual word similarity, we also investigate the effectiveness of our mapped cross-topic embeddings on document and sentence level downstream NLP tasks: text classification and paraphrase identification. We report weighted-averaging precision, recall, F1-measure and accuracy performance metrics.

\noindent
\textbf{Text classification.} We  used  the  $20$NewsGroup\footnote{http://qwone.com/~jason/20Newsgroups/} dataset, which consists of about $20\,000$ documents. Our goal is to classify each document into one of the $20$ different newsgroups based on its content.

\noindent
\textbf{Paraphrase Identification.} For this task we aimed at identifying whether two given sentences can be considered paraphrases or not, using
the Microsoft Paraphrase dataset~\cite{paraphrase}.

\noindent
\textbf{Document and Sentence level representations.}

\noindent
Given a document or a sentence $D$, where $w_d$ corresponds to the $d$-th word in $D$, we extract its feature representation using three different ways: 
\begin{align}
\small
\begin{split}\label{metrics:avgc}
\mathrm{AvgC_D} ={} &\frac{1}{|D|}\sum\limits_{d=1}^{|D|} \sum\limits_{k=1}^{K} \mathrm{p}(k|D) x'_k(w_d),\end{split}\\\notag
\\
\small
\begin{split}\label{metrics:avg}
\mathrm{Avg_D}\ \ \ ={} &\frac{1}{|D|}\sum\limits_{d=1}^{|D|} \sum\limits_{k=1}^{K} \frac{1}{K}x'_k(w_d),
\end{split}\\\notag
\\
\small
\begin{split}\label{metrics:maxc}
\mathrm{MaxC_D} = {}&  \frac{1}{|D|}\sum\limits_{w=1}^{|D|}  x'_m(w_d) \\
\\
\mathrm{s.t.} \ \ \ \ m ={} & \argmax_{k=1,..,K} \{\mathrm{p}(k|D)\},
\end{split}
\end{align}
where $\mathrm{p}(k|D)$ denotes the posterior probability of topic $k$ returned by LDA  given as input the sentence/document $D$ and $x'_k(w_d)$ is the mapped representation of word $w_d$ for topic $k$. For the case of paraphrase identification, we extract a single feature vector for each sentence-pair via concatenating the features of the individual sentences.

After feature extraction, we train a linear Support Vector Classifier (SVM)~\cite{scikit-learn} using the proposed train/test sets for both tasks. We report the best results for each experimental configuration after tuning the SVM's penalty parameter of the error term using $500$-dimensional word embeddings.

\section{Results}
In Table~\ref{tab:scws_res} we compare our model (UTDSM) with our baseline (Global-DSM) and other state-of-the-art multi-prototype approaches for the contextual semantic similarity task. It is clear that all different setups of UTDSM perform better than the baseline for both contextual semantic similarity metrics. Using a single Gaussian distribution (UTDSM + GMM (1)) at the smoothing step of our method produces similar results to the baseline model. This is anticipated as both methods provide a centroid representation of a word's diverse semantics.
In terms of $\mathrm{MaxSimC}$ the model consistently yields higher performance when the list of semantic anchors is induced via our unsupervised method instead of using randomly selected anchor words (UTDSM Random). We also observe that random anchoring performs slightly worse than UTDSM with respect to $\mathrm{AvgSimC}$. 
This result validates our hypothesis that the representations of words, which share consistent similarity distributions across different topic domains, constitute informative \textit{semantic anchors} that determine the mappings between semantic vector spaces.

\begin{table}[!ht]
    
    \scalebox{0.94}{
    \begin{tabular}{|l||c |c |}
    \hline
     Method & AvgSimC & MaxSimC \\
    \hline
    \hline
    \citet{Liu2015cont} & $67.3$ & $68.1$  \\
    \citet{liu2015topical} & $69.5$ & $67.9$  \\
    \citet{amiri2016learning} & $70.9$ & - \\
    \citet{Lee2017MUSEMU} & $68.7$& $67.9$ \\
    \citet{gr}  & $69.3$ & $68.2$ \\

    \hline
    \multicolumn{3}{c}{\textit{300-dimensions}}\\
    \hline
    Global-DSM  & $67.1$ & $67.1$ \\
    UTDSM Random  & $69.1 \pm 0.1$  & $66.4 \pm 0.2$ \\
    UTDSM  & $\mathbf{69.6}$ & $67.1$  \\
    UTDSM + GMM (1) & $67.4$ & $67.4$  \\
    UTDSM + GMM (2) & $68.4$ & $\mathbf{68.3}$  \\
    UTDSM + GMM (3) & $68.9$ & $\mathbf{68.3}$  \\
    UTDSM + GMM (8) & $69.1$ & $68.0$  \\
    UTDSM + GMM (10) & $69.0$ & $67.8$  \\

    \hline
    \multicolumn{3}{c}{\textit{500-dimensions}}\\
    \hline
    Global-DSM & $67.6$ & $67.6$  \\
    UTDSM Random  & $69.4 \pm 0.1$ & $66.5 \pm 0.3$  \\
    UTDSM  & $\mathbf{70.2}$ & $68.0$  \\
    UTDSM + GMM (1) & $67.6$ & $67.6$  \\
    UTDSM + GMM (2) & $68.8$ & $\mathbf{68.6}$  \\
    UTDSM + GMM (3) & $69.0$ & $68.5$  \\
    UTDSM + GMM (8) & $69.5$ & $68.5$  \\
    UTDSM + GMM (10) & $69.2$ & $68.0$  \\
    \hline
    \end{tabular}}
    
    \caption{Performance comparison between different state-of-the-art approaches on SCWS, in terms of Spearman's correlation. UTDSM refers to the projected cross-topic representation, UTDSM Random refers to the case when random words served as anchors and GMM ($c$) corresponds to GMM smoothing with $c$ components.}
    \label{tab:scws_res}
\end{table}

Furthermore, we observe that GMM smoothing has a different effect on the $\mathrm{MaxSimC}$ and $\mathrm{AvgSimC}$ metrics. Specifically, for $\mathrm{AvgSimC}$ we consistently report lower results when GMM smoothing is applied for different number of components. We attribute this behavior to a possible loss of model capacity---decrease in the number of topic embeddings---that is capable of capturing additional topic information. At the same time, our smoothing technique highly improves the performance of $\mathrm{MaxSimC}$ for all possible configurations. Given that this metric is more sensitive to noisy word representations, this result indicates that our technique lessens the noise introduced to our system and captures finer-grained topic senses of words. 

Overall, the performance of our model is highly competitive to the state-of-the-art models in terms of $\mathrm{AvgSimC}$, for $500$-dimensional topic embeddings. We also achieve state-of-the-art performance for the $\mathrm{MaxSimC}$ metric, using smoothed topic embeddings of $300$ or $500$ dimensions with $2$ or $3$ Gaussian components. 

\begin{table}[!ht]
    \captionsetup{font=small}
    \scalebox{0.8}{
    \begin{tabular}{|l | cccc|}
    \hline
    Method & Precision & Recall & F1-score & Accuracy\\
    \hline
    \hline
    LDA  & $39.7$ & $41.8$ & $38.8$ &  $41.8$\\
    Global-DSM  & $62.9$ & $63.3$ & $62.9$ & $63.3$\\
    \hline
    $\mathrm{MaxC_D}$  & $61.9$ &  $63.0$ & $62.0$ & $63.0$\\
    $\mathrm{Avg_D}$ & $63.5$ & $64.6$ & $63.3$ & $64.3$\\
    $\mathrm{AvgC_D}$ & $\mathbf{64.6}$ & $
    \mathbf{65.5}$ &  $\mathbf{64.5}$ & $\mathbf{65.5}$\\
    \hline
    \end{tabular}}
    \caption{Evaluation results of multi-class text classification.}
    \label{tab:newsgroup}
\end{table}

Evaluation results on text classification are presented in Table~\ref{tab:newsgroup}. We observe that our model performs better than the baseline across all metrics for both averaging approaches ($\mathrm{AvgC_D}$, $\mathrm{Avg_D}$), while the usage of dominant topics appears to have lower performance ($\mathrm{MaxC_D}$). Specifically, we get an improvement of $2-2.5\%$ on topic-based average and $0.5-1\%$ on simple average combination compared to using Global-DSM.
\begin{table}[!ht]
    \captionsetup{font=small}
    \scalebox{0.8}{
    \begin{tabular}{|l | cccc|}
    \hline
    Method & Precision & Recall & F1-score & Accuracy\\
    \hline
    \hline
    Global-DSM  & $68.6$ & $69.2$ & $62.0$  & $69.2$\\
    \hline
    $\mathrm{MaxC_D}$  & $\mathbf{69.0}$ & $69.3$ & $62.1$ & $69.3$ \\
    $\mathrm{Avg_D}$ & $67.7$ &  $\mathbf{69.4}$  & $\mathbf{64.0}$ & $\mathbf{69.4}$ \\
    $\mathrm{AvgC_D}$ & $68.8$ & $69.4$ & $62.6$ & $69.4$\\
    \hline
    \end{tabular}}
    \caption{Evaluation results on paraphrase detection task.}
    \label{tab:paraphrase}
\end{table}

Results for the paraphrase identification task are presented in Table~\ref{tab:paraphrase}. $\mathrm{Avg_D}$ yields the best results especially in F1 metric showing that cross-topic representations are semantically richer than single embeddings baseline (Global-DSM). Although we apply the topic distributions $\mathrm{p}(k|D)$ extracted from LDA (document-level model) to a sentence-level task, improvements over the baseline are also shown in the $\mathrm{AvgC_D}$ and $\mathrm{MaxC_D}$ cases. 

Overall, the proposed UTDSM model outperforms the baseline Global-DSM model on both semantic similarity and downstream tasks.\footnote{Similar results were obtained for each metric using smoothed word embeddings. Also, there are no standard evaluation approaches for comparison of previous works on downstream tasks.}
{
\renewcommand{\arraystretch}{1.15}
\begin{table*}[!ht]
    \captionsetup{font=small}
    \scalebox{0.8}{
    \begin{tabular}{l l l c}
    \toprule[1.4pt]
   
    \textbf{Word} & \textbf{Topic Words} &  \textbf{Nearest Neighbors} & \textbf{Similarity} \\
    \hline

    \multirow{2}{*}{drug} & health, medical, cancer, treatment, disease & insulin, therapy, heparin, chemotherapy, vaccines & \multirow{2}{*}{$0.61$}\\
    & drug, health, marijuana, alcohol, effects
 & meth, cocaine, methamphetamine, mdma, heroin & \\
    \hline
    
      \multirow{2}{*}{act} & law, court, legal, tax, state & bylaw, legislature, complying, entities, entitlement & \multirow{2}{*}{0.39}\\
    & music, guitar, piano, dance, theatre
 & touring, pantomime, weekend,  shakespeare, musical & \\
    \hline
    
    \multirow{2}{*}{python} & garden, plant, fish, bird, animal & macaw, crocodile, hamster, albino, rattlesnake
 & \multirow{2}{*}{$0.27$}\\
    & software, forum, download, windows, web 
 & algorithm, parser, notepad, gui, tutorial & \\
    \hline
    
  \multirow{2}{*}{rock} & mountain, river, park, road, trail
 & geology, slab, limestone, waterfalls, canyon & \multirow{2}{*}{$0.43$}\\
    & music, guitar, piano, dance, theatre
 & touring, acoustic, americana, songwriter, combo & \\
    \hline
  
    \multirow{2}{*}{nursery} & garden, plant, tree, flower, gardening & camellias, succulents, greenhouse, ornamental, grower & \multirow{2}{*}{$0.46$}\\
    & university, school, college, education, program
 & prep, montessori, grammar, preschool, infant & \\
    \hline
      
     \toprule[1.5pt]
    \end{tabular}} 
    \caption{Examples of polysemous words and the change of meaning between different topic domains. First column lists the example target words. Second column includes the most probable words of the topic domains\footnotemark \ \ these words are assigned to. Each row corresponds to a different topic domain. Third column shows the nearest monosemous neighbors of the target word in the corresponding topic domain. The last column corresponds to the cosine similarity between the two topic representations of the target word.} 
  \label{examples}
\end{table*}
}
\begin{figure*}[ht!]
    \captionsetup{font=small}
  \includegraphics[scale=0.23]{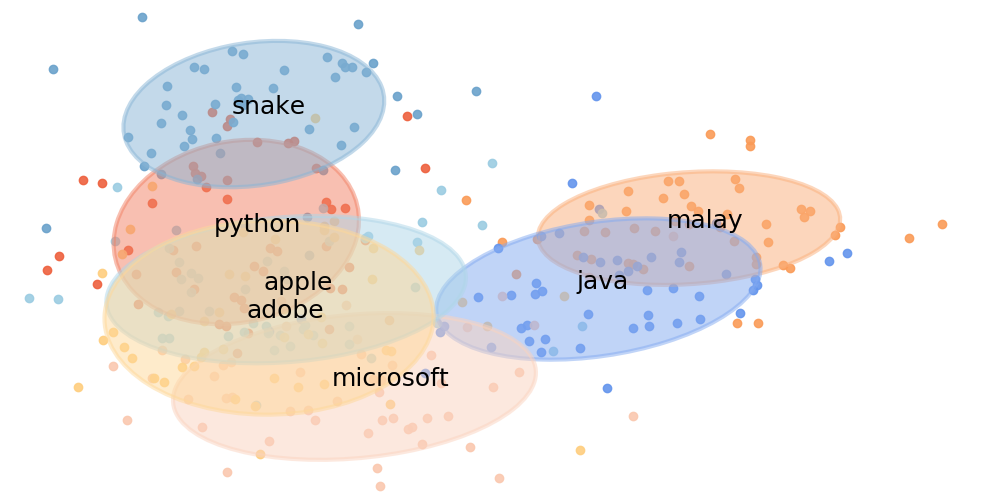}
  \includegraphics[scale=0.30]{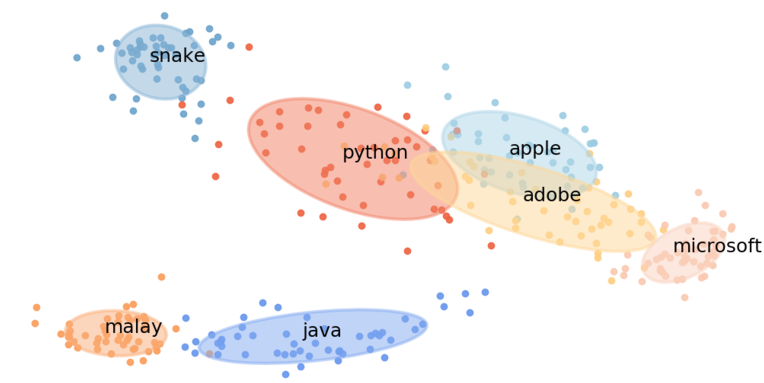}
   \caption{ A $2$-dimensional projection of the latent semantic space encoded in our unified vector space model, depicting the topic word representations of $7$ words before (left) and after (right) mapping the TDSMs to the global semantic space.}
    \label{python}
\end{figure*}

\section{Cross-domain semantic analysis}
Finally, we carry out a cross-domain semantic analysis to detect the variations of a word's meaning in different topic domains. To that end, we use a list of known polysemous words and measure the semantic similarity between different topic representations of the same ambiguous word. The ultimate goal of this analysis is to validate that our model captures known thematic variations in semantics of polysemous words.
\footnotetext{Note that a topic domain is described as a distribution over words in our model.}

Table \ref{examples} includes examples of our analysis. The most probable words of the topics (second column) give an intuitive sense of their major contexts, while their nearest neighbors (third column)
infer the sense of the target word in the corresponding topic domain. 
For example, the word \textit{drug} is mostly related to  ``medication'' in a broad medical domain; it experiences though a slight shift from this  meaning when it resides in a topic about ``illegal substances''. Furthermore, the highly polysemous word \textit{act} shifts from meaning ``statute'' to meaning ``performance'' under the corresponding law and art topics. Similar semantic variations are observed for words \textit{python}, \textit{rock} and \textit{nursery}.

Moreover, in Figure \ref{python} we visualize the topic embeddings of seven words before and after projecting the topic-based DSMs to the unified space, using principal component analysis. We additionally depict the Gaussian distribution learned from the topic representations of each word reflecting the uncertainty of their meanings. The center of each distribution is specified by the mean vector and contour surface by the covariance matrix.
On the left, we depict the position of words prior to applying the unsupervised mapping approach where the topic sub-spaces are unaligned. In the unaligned space, words demonstrate similar area coverage regardless of their polysemy. After the mappings, we see on the right that the area under a word's distribution is indicative of its degree of polysemy. Specifically, we observe that the variance of the learned representations becomes larger for the cases of polysemous words such as ``python'', ``java'', ``adobe''  in order to assign some probability to their diverse meanings. Monosemous words such as ``snake'', ``microsoft'' and ``malay'' have smaller variances. Furthermore, we observe that the semantic relationships between words are much better captured by their corresponding positions in the aligned space.

\section{Conclusion}

We present an unsupervised approach of mapping multiple topic-based DSMs to a unified vector space in order to capture different contextual semantics of words. We assume that words having consistent similarity distributions regardless of the domain they exist in could be considered informative semantic anchors that determine the mappings between semantic spaces. The projected word embeddings yield state-of-the-art results on contextual similarity compared to previously proposed unsupervised approaches for multiple word embeddings creation, while they also outperform single vector representations in downstream NLP tasks. In addition, we provide insightful visualizations and examples that demonstrate the capability of our model to capture variations in topic semantics of words. 

As future work, one can hypothesize that the area a word covers in the mapped space reveals its semantic range. In this direction, a refinement of the semantic anchor selection approach could be explored in an iterative way assuming that the variance of a word's Gaussian distribution denotes its degree of polysemy \cite{macallum}. Moreover, we would like to explore a more sophisticated smoothing technique where the number of Gaussian components is adapted for each word. Given that Gaussian mixture embeddings could capture the uncertainty of a word's representation in the semantic space one could also investigate different metrics for measuring the semantic relationship between word pairs that go beyond their point-wise comparison. Finally, it may be helpful to investigate non-linear mappings between semantic spaces using deep neural network architectures.
 
\section*{Acknowledgments}
Thi work has been partially funded by the BabyRobot project, supported by the EU Horizon $2020$ Program undergrant \#$687831$.

\bibliography{naaclhlt2019}
\bibliographystyle{acl_natbib}

\end{document}